\documentclass{article}
\pdfoutput=1
\usepackage{PRIMEarxiv}
\usepackage{amsmath,amssymb,amsfonts}
\usepackage{algorithmic}
\usepackage{algorithm}
\usepackage{graphicx}
\usepackage{booktabs}
\usepackage{tablefootnote}
\usepackage{threeparttable}
\usepackage[switch, pagewise]{lineno}  
\usepackage[authoryear]{natbib}
\usepackage{stfloats}  
\usepackage{hyperref}

\begin{document}
\title{Time-space-frequency feature Fusion for 3-channel motor imagery classification}

\author{
    Zhengqing Miao \footnotemark[1] \And
    Meirong Zhao   \footnotemark[1] \And
    }
\renewcommand{\thefootnote}{\fnsymbol{footnote}}
\footnotetext{Corresponding author: Zhengqing Miao (mzq@tju.edu.cn)}
\footnotetext[1]{State Key Laboratory of Precision Measuring Technology and Instruments, School of Precision Instrument and Opto-electronics Engineering, Tianjin University, Tianjin 300072, China.}

\maketitle

\begin{abstract}
Low-channel EEG devices are crucial for portable and entertainment applications. However, the low spatial resolution of EEG presents challenges in decoding low-channel motor imagery. This study introduces TSFF-Net, a novel network architecture that integrates time-space-frequency features, effectively compensating for the limitations of single-mode feature extraction networks based on time-series or time-frequency modalities.
TSFF-Net comprises four main components: time-frequency representation, time-frequency feature extraction, time-space feature extraction, and feature fusion and classification. Time-frequency representation and feature extraction transform raw EEG signals into time-frequency spectrograms and extract relevant features. The time-space network processes time-series EEG trials as input and extracts temporal-spatial features. Feature fusion employs MMD loss to constrain the distribution of time-frequency and time-space features in the Reproducing  Kernel Hilbert Space, subsequently combining these features using a weighted fusion approach to obtain effective time-space-frequency features.
Moreover, few studies have explored the decoding of three-channel motor imagery based on time-frequency spectrograms. This study proposes a shallow, lightweight decoding architecture (TSFF-img) based on time-frequency spectrograms and compares its classification performance in low-channel motor imagery with other methods using two publicly available datasets. Experimental results demonstrate that TSFF-Net not only compensates for the shortcomings of single-mode feature extraction networks in EEG decoding, but also outperforms other state-of-the-art methods. Overall, TSFF-Net offers considerable advantages in decoding low-channel motor imagery and provides valuable insights for algorithmically enhancing low-channel EEG decoding.


\end{abstract}

\section{Introduction}
\label{sec:introduction}

Electroencephalography (EEG) has become increasingly important in various fields such as motor rehabilitation, brain function regulation, and entertainment [\cite{park2014assessment, sterman1996physiological, olfers2018game}]. As a non-invasive technique, EEG measures brain activity by detecting and recording electrical signals produced by neurons firing in the brain [\cite{nunez2006electric, niedermeyer2005electroencephalography}]. The main advantages of EEG include its high temporal resolution, relatively low cost, and accessibility for a wide range of clinical and research applications[\cite{lotte2018review}]. In portable devices and entertainment applications, low-channel EEG acquisition systems play a crucial role, significantly enhancing the ease of use and accessibility of EEG-related devices for users [\cite{minguillon2017trends}].

Motor imagery is a common paradigm for EEG data collection, widely employed in brain-computer interfaces (BCIs) . During motor imagery, the sensorimotor cortex, which includes the primary motor cortex (M1) and primary somatosensory cortex (S1), generates neural oscillations that can be captured by EEG [\cite{pfurtscheller2001motor}]. These oscillations are categorized into different frequency bands, such as mu (8-12 Hz) and beta (13-30 Hz) rhythms. The power of these rhythms changes in response to motor imagery tasks, resulting in event-related desynchronization (ERD) or event-related synchronization (ERS) patterns[\cite{pfurtscheller2006mu}].

Common spatial pattern (CSP) is a widely used spatial filtering technique that maximizes the variance between two classes of motor imagery signals [\cite{lotte2010regularizing}]. It enables the extraction of discriminative features for classification [\cite{ramoser2000optimal}]. However, CSP-related methods have limited feature extraction capability and require a more stringent number of EEG channels.
\cite{baig2020filtering} state that in CSP-based MI classification studies, it is common to filter 10-30 EEG channels for feature extraction. 
\cite{arvaneh2011optimizing} also showed that using only three channels (C3, C4, and Cz) for left- and right-handed binary classification of motor imagery demonstrated a significant decrease in classification accuracy compared to using a selected 13 EEG channels. This makes the CSP-based feature extraction method very limited for MI feature extraction in low channels.

In recent years, researchers have increasingly explored the powerful feature extraction capability of artificial neural networks to automatically extract and classify effective features of MI signals [\cite{al2021deep,  craik2019deep}]. There are two major categories of such methods: one inputs time-series EEG signals into artificial neural networks for feature extraction and classification, and the other converts raw EEG into images first and then inputs them into artificial neural networks for feature extraction and classification.

Considering the obvious time-frequency characteristics of EEG, some works utilize the outstanding performance of convolutional neural networks (CNNs) in natural image feature extraction to extract and classify effective features in the spectrogram of EEG [\cite{mammone2020deep}]. For instance,
\cite{khare2020time} first transformed band-pass filtered EEG signals into time-frequency representations using smoothed pseudo-Wigner-Ville distribution (SPWVD), and then extracted features using classical CNN architectures such as pretrained AlexNet [\cite{krizhevsky2017imagenet}], VGG[\cite{simonyan2014very}], and ResNet[\cite{he2016deep}], achieving significant results in emotion recognition. 
\cite{madhavan2019time} employed Fourier synchrosqueezing transform and wavelet synchrosqueezing transform to convert raw EEG into time-frequency matrices, and then used a deep CNN network for feature extraction, yielding good performance in the classification of focal and non-focal EEG signals. 
\cite{xu2020learning} first converted raw EEG into EEG topographical representations, then employed convolutional neural networks for feature extraction, which performed well in motor imagery classification.

In contrast to transforming raw EEG into images first, neural network decoding methods that use time-series EEG signals as input eliminate the need for an additional transformation step. 
\cite{schirrmeister2017deep} presented a representative work that directly inputs time-series EEG signals into artificial neural networks for feature extraction. By setting appropriate time convolutional layers and spatial convolutional layers, they demonstrated that end-to-end convolutional neural networks can achieve feature extraction capability comparable to that of filter bank common spatial patterns (FBCSP) [\cite{ang2008filter}]. EEGNet [\cite{lawhern2018eegnet}] is another compact CNN architecture with time-series EEG signals as inputs, which exhibits excellent classification performance in four BCI paradigms, namely P300 visual-evoked potentials, error-related negativity, movement-related cortical potentials, and sensory motor rhythms.
\cite{miao2023lmda} introduces a lightweight network architecture based on channel attention and depth attention, which performs well in feature extraction and classification of motor imagery and error-related negativity with different EEG channels. 
\cite{borra2020interpretable} proposed a lightweight network framework for motor imagery feature extraction and classification, extracting effective features of motor imagination signals through a temporal sinc-convolutional layer and a spatial depthwise convolution. 

However, among the existing decoding methods, there is no motor imagery decoding algorithm specifically designed for low EEG channel scenarios. One possible reason is the low spatial resolution of EEG, leading many studies to attempt to improve the accuracy of motor imagery classification by increasing the number of electrodes [\cite{arvaneh2011optimizing, baig2020filtering}]. 
Nonetheless, in areas such as portable EEG devices and entertainment, increasing the number of EEG channels can be inconvenient to use, making it significant to improve the classification of low channel count EEGs from an algorithmic perspective. Therefore, this study combines the advantages of time-series neural networks and image-based neural networks to enhance the classification accuracy of low channel count motor imagery from an algorithmic perspective, promoting the application of EEG in portable devices and entertainment.

In this study, we propose a  time-space-frequency feature fusion method to extract and classify motor imagery features in three-channel scenarios using the proposed time-space-frequency fusion network (TSFF-Net). TSFF-Net consists of four components, namely time-frequency representation, spectral-based time-frequency feature extraction layers, time-series-based time-space feature extraction layers, and feature fusion and classification. Results on two motor imagery public datasets show that TSFF-Net not only outperforms the state-of-the-art methods but also surpasses the 22-channel methods by using only three EEG channels. In a word, TSFF-Net holds great potential for decoding low EEG channels.

The major contributions of this paper can be summarized as follows:

\begin{enumerate}
\item A network architecture for three-channel MI-EEG decoding is proposed from the perspective of multimodal fusion, effectively fusing time-space-frequency features to compensate for the shortcomings of single-modal networks in feature extraction for 3-channel MI-EEG.
\item A three-channel MI-EEG spectral feature extraction network architecture, TSFF-Img, is proposed, which is more advantageous in decoding three-channel MI-EEG compared with natural image network architectures such as AlexNet, VGG, and ResNet.
\item Comprehensive experiments on two public motor imagery datasets validate the effectiveness of the proposed method, providing a valuable reference for future research on low-channel EEG decoding.
\end{enumerate}

The remainder of this paper is organized as follows: Section \ref{sec:method} introduces the proposed time-space-frequency feature fusion network (TSFF-Net) in detail. In Section \ref{sec:ExperimentSettings}, we present the datasets, experimental setup, and preprocessing methods. Section \ref{sec:Results} discusses results and the key findings of this study. Finally, Section \ref{sec:Conclusion} concludes the paper.

\begin{figure*}
\centerline{\includegraphics[width=\textwidth]{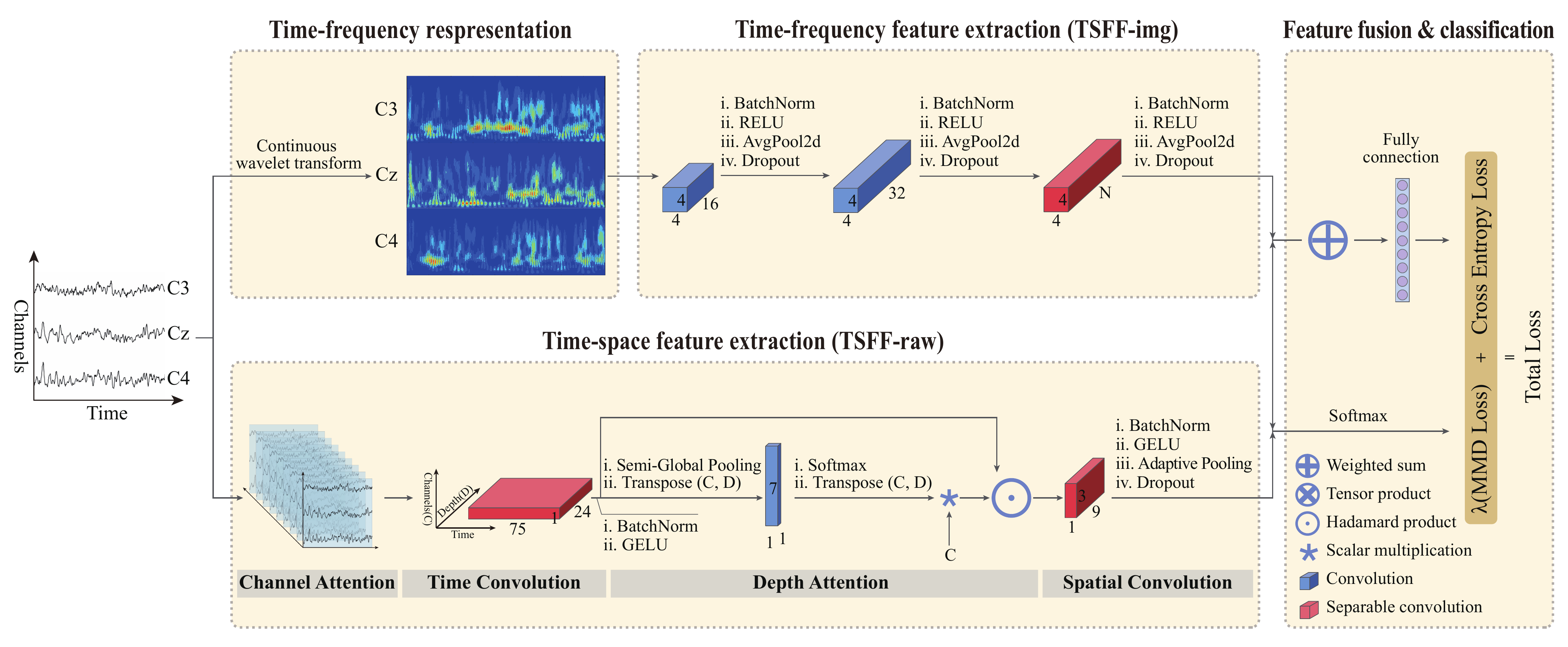}}
\caption{Time-space-frequency feature fusion network architecture (TSFF-Net). TSFF-Net contains four components, namely time-frequency representation, time-frequency feature extraction, time-space feature extraction, and feature fusion and classification.}
\label{fig:tsff}
\end{figure*}

\section{Method}
\label{sec:method}

TSFF-Net takes the pre-processed segmented EEG trials as input and contains four parts: time-frequency representation, time-frequency feature extraction, time-space feature extraction, and feature fusion and classification (as shown in Figure \ref{fig:tsff}). Let the segmented EEG trials be $\mathbf{X} = \{\mathbf{x}_i\}_{i=1}^{N}$, $\mathbf{x}_i \in \mathbb{R}^{C\times{T}}$, where $N$ denotes the total number of training samples, $C$ denotes the number of EEG channels, and $T$ denotes the number of samples contained in each trial. The following four components of TSFF-Net are described in detail below.

\subsection{Time-frequency representation}
For each channel in $\mathbf{x}_i$, the temporal data are transformed into the corresponding frequency spectrum using the continuous wavelet transform (CWT)[\cite{rioul1992fast}]. Let $\mathbf{x}_{i,j}$ denote the temporal EEG data of the $j$th channel in $\mathbf{x}_i$, abbreviated as $s(t)$. The continuous wavelet transform is performed using the following equation:
\begin{equation}
CWT(a, b)=\frac{1}{\sqrt{a}} \int s(t) \psi^{}\left(\frac{t-b}{a}\right) dt
\end{equation}
where $\psi$ represents the wavelet mother function, $\psi^{}$ is the complex conjugate of $\psi$, $a$ and $b$ denote the dilatation and shifting variables. In this study, Morlet wavelet [\cite{lin2000feature}] is selected as the wavelet mother function, which is defined as
\begin{equation}
\psi(t)=\exp \left(-\frac{\beta^{2} t^{2}}{2}\right) \cos (\pi t)
\end{equation}
where $\beta$ is a parameter to balance the time resolution and frequency resolution of the Morlet wavelet.

The output of the CWT is a complex-valued matrix, where each element represents the wavelet coefficient at a specific time and frequency. The magnitude of the wavelet coefficient represents the strength of the EEG activity at that time and frequency. The frequency range of the EEG spectrogram is aligned with the passband of the original EEG bandpass filter to reduce the effect of noise on the spectrogram. To generate a time-frequency map, the magnitude of the wavelet coefficients is squared and then plotted as a function of time and frequency.
CWT of each EEG segment yields 800 x 600 time-frequency spectrum image. In TSFF-Net, the frequency spectrum images corresponding to the C3, Cz, and C4 channels are stitched together in the widthwise direction, and the stitched images are downsampled to 224 x 224 and then fed into the frequency-spectrum feature extraction part of TSFF-Net for feature extraction.

\subsection{Time-frequency feature extraction}

\begin{table*}[]
\caption{TSFF-img network architecture and number of learnable parameters}
\centering
\label{tabel:tsff-img}
\begin{tabular}{ccccccccc}
\hline
Layer(l) & Layer type      & \#Filters & Filter size & Stride & padding & groups & Bias & \# Parameters \\ \hline
c1       & Convolution     & 16        & 4×4         & 1      & 2       & 1      & 16   & 784           \\
r1       & ReLU            & -         & -           & -      & -       & -      & -    & -             \\
r1       & AvgPool         & 16        & 8×8         & -      & -       & -      & -    & -             \\
r1       & Dropout         & -         & P=0.25      & -      & -       & -      & -    & -             \\
c2       & Convolution     & 32        & 4×4         & 1      & 2       & 1      & 32   & 8224          \\
r2       & ReLU            & -         & -           & -      & -       & -      & -    & -             \\
r2       & AvgPool         & 32        & 3×3         & -      & -       & -      & -    & -             \\
r2       & Dropout         & -         & P=0.25      & -      & -       & -      & -    & -             \\
c3       & Convolution     & 64        & 1×1         & 1      & 0       & 1      & 0    & 2048          \\
r3       & BatchNorm       & 64        & -           & -      & -       & -      & -    & 128           \\
c3       & Convolution     & 64        & 4×4         & 1      & 2       & 64     & 0    & 1024          \\
r3       & BatchNorm       & 64        & -           & -      & -       & -      & -    & 128           \\
r3       & ReLU            & -         & -           & -      & -       & -      & -    & -             \\
r3       & AvgPool         & 32        & 3×3         & -      & -       & -      & -    & 0             \\
r3       & Dropout         & -         & P=0.25      & -      & -       & -      & -    & 0             \\
fc       & Fully Connected & -         & -           & -      & -       & -      & 2    & 1154          \\ \hline
Total    &                 &           &             &        &         &        &      & 13490         \\ \hline
\end{tabular}
\end{table*}
As shown in Table \ref{tabel:tsff-img}, the proposed TSFF-img network comprises three feature extraction layers, each followed by a series of regularization operations. These regularization operations are consistent with those used in the raw EEG feature extraction network. In contrast to the maximum pooling operation commonly employed in natural image feature extraction network architectures, the pooling layer in the TSFF-img network uses average pooling due to the low signal-to-noise ratio of EEG signals and their susceptibility to noise contamination. The parameters of the feature extraction layers are fixed in the first two layers of the frequency-spectrum network, utilizing kernels of (16, 4, 4) and (32, 4, 4), where 16 and 32 represent the number of convolution kernels and (4, 4) denotes the size of the convolution kernel. To reduce the number of parameters in TSFF-img, the final convolution layer employs separable convolution. Table \ref{tabel:tsff-img} displays the number of parameters in TSFF-img when the input segmented EEG trial has a length of 1000. The number of kernels used in the last convolutional layer is determined based on the output of the time-space feature extraction layers (TSFF-raw), ensuring that the dimensionality of the features output from the TSFF-img network matches that of the features output from the TSFF-raw for subsequent feature fusion. 

As shown in Table \ref{tabel:tsff-img}, when the input segmented EEG trial has a time length of 1000, TSFF-img has only 13,490 learnable parameters for the binary classification task. This is substantially fewer than classical natural image recognition networks such as AlexNet [\cite{krizhevsky2017imagenet}], VGG[\cite{simonyan2014very}], and ResNet[\cite{he2016deep}], which possess millions or billions of parameters. TSFF-img effectively extracts features with a small number of parameters and a limited dataset size. Additionally, the TSFF-img network consists of only three layers, which aim to learn shallow information, such as contours in the EEG time-frequency spectrum, while ignoring detailed information in the time-frequency spectrum to reduce noise interference.

\subsection{Time-space feature extraction}

The time-space feature extraction layers used in this study is the same as the feature extraction employed in LMDA-Net [\cite{miao2023lmda}]. 
The feature extraction layer in LMDA-Net is divided into four components: channel attention model, time convolutional layer, depth attention model, and spatial convolutional layer (as shown in Figure \ref{fig:tsff}). LMDA-Net is also a lightweight network architecture, employing separable convolution in both time convolutional and spatial convolutional layers.

In LMDA-Net, the channel attention module diffuses the spatial information of the EEG signal to the depth dimension through tensor product operations, mitigating the problem of low spatial resolution in EEG signals due to volume conduction during acquisition. Let $\mathbf{X}$ be the input sample of time-series EEG and $\mathbf{C}$, $\mathbf{C} \in \mathbb{R}^{D\times {1} \times {C}}$, be a tensor following a normal distribution. The channel attention module equation is given as follows:
\begin{equation}
    \mathbf{X}_{h c t}^{\prime}=\sum_{d} \mathbf{X}_{d c t} \mathbf{C}_{h d c} \label{eq3}
\end{equation}
The subscripts in equation (\ref{eq3}) indicate the corresponding dimensions. Matching letter subscripts signify that the two tensors share the same shape in that dimension.

LMDA-Net features a depth attention module between the time convolution and spatial convolution layers (as shown in Figure \ref{fig:tsff}). The depth attention module strengthens the connection between time-domain and spatial-domain features of time-series EEG signals, enabling the full exploitation of temporal and spatial features of these signals.

\subsection{Feature fusion and classification}
The feature fusion in TSFF-Net comprises two components: weighted fusion of the output features from the time-space and time-frequency feature extraction layers, and alignment of these output features in high-dimensional space through MMD loss [\cite{gretton2012kernel}]. Since time-space and time-frequency features are derived from different representations of the same type of EEG signal, they should exhibit similar feature distributions in certain high-dimensional spaces. Both feature-weighted fusion and MMD loss ensure that features extracted by neural networks of different modalities can be effectively combined to address the limitations of single-modal feature extraction networks.

Let the flattened output features of TSFF-raw be $\mathbf{S}$ and those of TSFF-img be $\mathbf{F}$, with ($\mathbf{F}, \mathbf{S} \in \mathbb{R}^{N\times{D}}$). Due to the amplitude mismatch between time-frequency and time-space features, $\mathbf{S}$ and $\mathbf{F}$ are normalized using the softmax function before applying MMD loss. For instance, for $\mathbf{S}$, the softmax function is expressed as:
\begin{equation}
 \mathbf{S}(x_i) = \frac{\exp \left(x_i\right)}{\sum_{j=1}^N \exp \left(x_{j}\right)}
\end{equation}
where $x_i$ denotes the flattened features of the $i$th sample in $\mathbf{S}$. 

The normalized $\mathbf{S}$ and $\mathbf{F}$ are then input to the MMD loss, defined as: 
\begin{equation}
d_{k}^{2}(p, q) \triangleq\left\|\mathbf{E}_{p}\left[\phi\left(\mathbf{S}\right)\right]-\mathbf{E}_{q}\left[\phi\left(\mathbf{F}\right)\right]\right\|_{\mathcal{H}_{k}}^{2}
\end{equation}
where, $p$ and $q$ represent the probability distributions of $\mathbf{S}$ and $\mathbf{F}$, respectively. $\mathbf{E}_{p}$ and $\mathbf{E}_{q}$ denote the mean embeddings of probability distributions $p$ and $q$. $\phi(\cdot)$ signifies the kernel function of feature mapping, while the subscript $\mathcal{H}_{k}$ indicates that the mean difference of the two feature distributions should be constrained to a unit ball in the Reproducing Kernel Hilbert Space (RKHS) [\cite{rosipal2001kernel}].

The input of the fully connected layer in TSFF-Net is the weighted average of $\mathbf{S}$ and $\mathbf{F}$. Let $\eta$, with $0 \leq \eta \leq 1$, represent the feature weight, and the weighted fusion of time-frequency and time-space network output features can be expressed as:
\begin{equation}
    \mathbf{G}=\eta \cdot \mathbf{S}+(1-\eta) \cdot \mathbf{F} 
\end{equation}

The fused feature $\mathbf{G}$ is then input to the fully connected network and classified using CrossEntropy Loss [\cite{shannon2001mathematical}]. Let the prediction result of TSFF-Net be $P$, and $Y$ represent the true label of the predicted sample. The CrossEntropy Loss can be expressed as:
\begin{equation}
L_c(Y, P)=-\frac{1}{N} \sum_{i=1}^{N} \sum_{m=1}^{M} y_{i, m} \log \left(p_{i, m}\right)
\end{equation}
where $N$ denotes the number of training samples, $M$ represents the number of motion imagery categories, and $m$ corresponds to a specific motion imagery category. $y_{i,m}$ denotes the $m$th element of the true label $Y$, while $p_{i,m}$ represents the probability value that the corresponding sample is predicted as category $m$ in TSFF-Net.

During the training process of TSFF-Net, the total loss function is expressed as:
\begin{equation}
L = L_c + \lambda L_{MMD}
\end{equation}
where $\lambda$ represents the weight of MMD loss.

\section{Experiment settings}
\label{sec:ExperimentSettings}

\subsection{Datasets}
\subsubsection{BCI Competition IV Dataset IIA (BCI4-2A)}
The BCI4-2A dataset \footnote{\url{www.bbci.de/competition/iv/\#dataset2a}} consists of EEG data collected using a 10-20 system with 22 EEG channels at a sampling rate of 250 Hz from nine healthy participants (ID A01-A09) across two different sessions. Each participant performed four motor imagery tasks, including imagining the movement of the left hand, right hand, both feet, and tongue. Each session contains 288 trials of EEG data. All data collected in the first session were used for training, while the second session data were used for testing. A temporal segmentation of [2, 6] seconds after the MI cue was extracted as one trial of EEG data in our experiment. In addition to the quadruple classification task, a separate motor imagery task for left- and right-handed binary classification was studied, with 144 trials per session. The dataset contained 22 channels; however, in subsequent experiments, we selected only the data corresponding to channels C3, Cz, and C4 for analysis.

\subsubsection{BCI Competition IV Dataset IIB (BCI4-2B)}
The BCI4-2B dataset \footnote{\url{www.bbci.de/competition/iv/\#dataset2b}} comprises EEG data collected using 3 EEG electrode channels (C3, Cz, and C4) sampled at 250 Hz from nine healthy participants (ID B01-B09) over five separate acquisition sessions. Each participant performed two motor imagery tasks, which involved imagining the movement of the left hand and right hand. The first two sessions contained 120 trials per session without feedback, while the last three sessions included 160 trials per session with a smiley face on the screen as feedback. All data from the first three sessions were used for training, and the last two sessions were used for testing. A temporal segmentation of [3, 7] seconds after the MI cue was extracted as one trial of EEG data in our experiment.

\subsection{Preprocessing}
A 200-order Blackman windows bandpass filter was used to filter the raw EEG data. The passband of the filter was chosen [4, 38] Hz. Then the filtered raw EEG data was segmented according to the task duration in each dataset. The segmented EEG trials are fed into the network framework in Figure \ref{fig:tsff} for training after the following two pre-processing steps in turn.

A 200-order Blackman window bandpass filter was employed to preprocess the raw EEG data. The passband of the filter was selected to be within the range of [4, 38] Hz, effectively capturing the primary frequency bands of interest associated with motor imagery tasks, such as mu (8-12 Hz) and beta (13-30 Hz) rhythms. Subsequently, the filtered raw EEG data was segmented according to the specific task duration pertinent to each dataset, extracting individual trials for further analysis. The segmented EEG trials underwent the following two sequential preprocessing steps before being input into the TSFF-Net.

\subsubsection{Normalization}
To eliminate the impact of data scale on neural network training and improve the convergence speed of the network, we first normalized the amplitudes obtained for each trial. The EEG data of each trial were normalized to [-1, 1] by the following equation.

Considering that this study focuses on a limited number of channels, specifically three, channels C3 and C4 may hold more significant weight than channel Cz, thus requiring a tailored normalization approach that differs from the channel normalization employed in  [\cite{miao2022priming}]. We adopted a trial-based normalization strategy to preprocess the EEG data. This method ensures that the amplitude values obtained for each trial are consistently normalized, thus mitigating any biases that may arise due to channel-specific characteristics. The EEG data for each trial were scaled to the range of [-1, 1] using the following equation:
\begin{equation}
    \mathbf{x}_{i}=\frac{\mathbf{x}_{i}}{\max \left(\left|\mathbf{x}_{i}\right|\right)}
\end{equation}
Where i denotes the $i$th trial of $\mathbf{x}$, $\left|\cdot\right|$ denotes taking the absolute value of the matrix.

\subsubsection{Euclidean alignment}

In order to ensure the robustness and generalizability of our feature extraction process, we adopted the domain alignment methodology from  [\cite{miao2022priming}] to construct a domain-invariant representation of both training and test data. This approach aims to establish a consistent and unified mean covariance matrix across all n aligned trials, effectively minimizing the distribution discrepancy between the training and test data sets. This process, which results in a unit matrix [\cite{he2019transfer}], ultimately facilitates more accurate and reliable feature extraction.

The specific implementation of this Euclidean alignment methodology is as follows:
\begin{align}
\bar{\mathbf{R}}&=\frac{1}{N} \sum_{i=1}^{N} \mathbf{x}_{i} \mathbf{x}_{i}^{T} \notag\\
\tilde{\mathbf{x}}_{i}&=\bar{\mathbf{R}}^{-1 / 2} \mathbf{x}_{i}
\end{align}
By incorporating the domain alignment technique into our study, we are effectively addressing potential discrepancies between the training and test data that may arise due to inherent differences in their respective distributions.

\subsection{Experimental environment and parameter settings}
All experiments were conducted using the PyTorch (Version 1.10.1) framework on a high-performance workstation equipped with Intel(R) Xeon(R) Gold 5117 CPUs operating at 2.00 GHz, and Nvidia Tesla V100 GPUs. To ensure optimal performance, we employed the AdamW optimizer for training all models, utilizing the default parameters as described by \cite{loshchilov2018fixing}. The training process was conducted using mini-batches of size 32. And all networks were trained for no more than 350 epochs.     

It is important to note that the ConvNet architecture proposed by \cite{schirrmeister2017deep} 
has both shallow and deep versions for EEG decoding. However, previous research, including works by \cite{lawhern2018eegnet} 
and \cite{schirrmeister2017deep} themselves, have suggested that the shallow-ConvNet offers more advantages compared to the deep-ConvNet in the context of EEG decoding. Consequently, we opted to utilize the shallow-ConvNet architecture (hereafter referred to as ConvNet) in our experiments.

\section{Results}
\label{sec:Results}

In Section \ref{sec:results analysis}, we assessed the best classification performance of various methods in the BCI4-2A binary and quadruple classification tasks, as well as the BCI4-2B binary classification task.In Section \ref{sec:ANNs and CSP}, we compared the prediction performance of several neural network models on the binary classification task of the BCI4-2A dataset. Specifically, we evaluated the average accuracy of all participants for each epoch of EEGNet, ConvNet, LMDA-Net, TSff-img, and TSFF-Net, and compared their performance with models based on CSP. In Section \ref{sec:spectrogram models}, we compared the classification performance of the proposed TSFF-img with pre-trained AlexNet, VGG, and ResNet in the EEG spectrogram. In Section \ref{sec:ablation}, we conducted ablation experiments to demonstrate the impact of various modules on TSFF-Net. 

\subsection{Experimental results analysis} \label{sec:results analysis}

\subsubsection{Binary classification of BCI4-2A}
We first analyzed the classification performance of different models in BCI4-2A for the left- and right-handed dichotomous classification task.  Since BCI4-2A is a quadruple classification dataset, neural network related work generally demonstrates the performance when multiple classifications are performed. To this end, we replicated the classification performance of these three network models when using only the C3, Cz, and C4 channels, based on the open source codes of EEGNet [\cite{lawhern2018eegnet}], ConvNet[\cite{schirrmeister2017deep}, and LMDA-Net [\cite{miao2023lmda}]. EA-CSP[\cite{he2019transfer}] and SCSP[\cite{arvaneh2011optimizing}] are used to focus on the classification performance of the proposed method in BCI4-2A left- and right-handed binary classification. 
Among them, SCSP also explores the classification performance of the proposed method when only C3, Cz and C4 channels are used. Moreover, in the binary classification task of BCI4-2A, SCSP achieved same average accuracy to \cite{zanini2017transfer} with only three channels, whereas the latter used 22 channels. Thus, the classification performance of SCSP can serve as a representative method for CSP-related approaches.
Some of the preprocessing methods of neural networks are partly based on the idea of aligning the mean covariance matrices between participants in Euclidean space in \cite{he2019transfer}, so we also compare the classification performance of the neural network correlation method with EA-CSP for 2-classification. TSFF-img is a lightweight spectral analysis method proposed in this paper, and we also tested the difference in classification performance between it and the time-series-based neural network feature extraction methods in the same scenario. 

From Table \ref{tabel:2a2cls}, we can see that using only three channels, SCSP achieved an average accuracy of 70.9\%, which improved to 79.2\% when 22 channels were utilized. In comparison, EA-CSP achieved an average classification accuracy of 73.5\% using 22 channels.
When comparing the classification performance of the neural network methods in Table \ref{tabel:2a2cls} with that of SCSP(3channels), the neural network methods dominate both in terms of mean accuracy and standard deviation. This indicates, to some extent, that the neural network approach is more advantageous than the CSP-related approach in the dichotomous performance of motor imagery at low channel counts. However, LMDA-Net, ConvNet and TSFF-Net are ahead of SCSP(22channels) in terms of average accuracy and standard deviation , with using only 3 channels. 
Among the tested methods, the time-series-based models (i.e., LMDA-Net, ConvNet, and EEGNet) demonstrated superior performance compared to the time-frequency-spectrum-based approach represented by TSFF-img. However, the proposed method, TSFF-Net, can effectively improve the classification performance of both TSFF-img and TSFF-raw (i.e., LMDA-Net), which shows the effectiveness of the proposed fusion method in the low-channel binary classification MI scenario. 

\begin{table*}[]
\caption{Binary classification performance of different algorithms on BCI4-2A}
\centering
\label{tabel:2a2cls}
\begin{threeparttable}
\begin{tabular}{ccccccccccccc}
\hline
                      \textbf{Methods} & \textbf{A01}  & \textbf{A02}  & \textbf{A03}   & \textbf{A04}  & \textbf{A05}  & \textbf{A06}  & \textbf{A07}  & \textbf{A08}  & \textbf{A09}  & \textbf{Mean}  & \textbf{Std} & \textbf{P-Value} \\ \hline
EA-CSP(C=22) & 87.5          & 56.3          & 98.6           & 73.6          & 50.0          & 64.6          & 68.8          & 89.6          & 72.9          & 73.5          & 16.0         & 0.039            \\
SCSP(C=22)  & \textbf{91.0} & 56.3          & 96.5           & 72.9          & 63.9          & 63.9          & 79.9          & \textbf{97.2} & 91.7          & 79.2          & 15.6         & 0.12             \\ 
SCSP   & 75.7          & 53.5          & 93.1           & 68.1          & 53.5          & 61.1          & 57.6          & 86.8          & 88.9          & 70.9          & 15.7         & 0.011            \\ 
LMDA-Net                & 80.6          & 69.4          & 97.9           & 69.4          & 76.4          & 66.7          & \textbf{95.8} & 84.7          & 91.0          & 81.3          & 11.8         & 0.011            \\
ConvNet                & 86.8          & 67.4          & 96.5           & 68.1          & \textbf{79.9} & 61.1          & \textbf{95.8} & 83.3          & \textbf{92.4} & 81.2          & 13.1         & 0.062            \\
EEGNet                 & 82.6          & 63.9          & 97.2           & 68.1          & 73.6          & 65.3          & 82.6          & 84.7          & 91.0          & 78.8          & 11.7         & 0.0039           \\ 
\hline
TSFF-img         & 79.9          & 63.9          & 93.1           & 61.8          & 73.6          & 57.6          & 81.3          & 79.2          & 91.0          & 75.7          & 12.5         & 0.0039           \\ 
TSFF-Net        & 86.8          & \textbf{75.7} & \textbf{100.0} & \textbf{75.0} & 79.2          & \textbf{75.0} & \textbf{95.8} & 86.8          & 91.7          & \textbf{85.1} & 9.5          & -                \\ \hline
\end{tabular}
    \begin{tablenotes}
        \footnotesize
        \item P-value denotes the Wilcoxon signed-rank test between TSFF-Net and other results.
        \item The notation C=22 suggests that the method employs 22 channels, whereas the Others method uses only three channels (i.e., C3, C4, and Cz).
        \item The results of EEGNet, ConvNet, and LMDA-Net were obtained using their open-source network architectures, with an experimental setup identical to that of TSFF-Net.
    \end{tablenotes}
\end{threeparttable}
\end{table*}

\subsubsection{Binary classification of BCI4-2B}
We further validated the classification performance of different methods in BCI4-2B. FBCSP [\cite{ang2008filter}] is a representative method of manual feature extraction that enhance the feature extraction ability of CSP in multiple frequency bands by setting filter banks. WasF-ConvNet [\cite{zhao2019learning}] is a joint space-time-frequency feature extraction neural network for MI-EEG decoding. None of the above neural network methods employed data augmentation or training with the aid of other participants' data.

As shown in Table \ref{tabel:3}, the average accuracy of time-series based methods surpassed that of CSP-related methods on BCI4-2B, with the exception of ConvNet. In terms of accuracy, TSFF-img was not superior compared to the other neural network methods in Table \ref{tabel:3}. This may be attributed to the fact that time-series contains more information compared to time-frequency spectrum. However, TSFF-Net proposed in this study compensates for the shortcomings by fusing time-series features with time-frequency features to some extent.

It is worth mentioning that WasF-ConvNet [\cite{zhao2019learning}] is a method that combines time-space and frequency features, achieving the purpose of joint learning of time-space-frequency features by implementing time-frequency convolution and spatial convolution in turn. From the experimental results, the proposed method outperforms WasF-ConvNet in most participants in BCI4-2B and is 3.8\% ahead of WasF-ConvNet in terms of average accuracy. This demonstrates the feasibility of feature fusion.

\begin{table*}[]
\caption{Binary classification performance of different algorithms on BCI4-2B}
\centering
\label{tabel:3}
\begin{threeparttable}
\begin{tabular}{ccccccccccccc}
\hline
\textbf{Methods}       & \textbf{B01}  & \textbf{B02}  & \textbf{B03}  & \textbf{B04}  & \textbf{B05}  & \textbf{B06}  & \textbf{B07}  & \textbf{B08}  & \textbf{B09}  & \textbf{Mean}  & \textbf{Std} & \textbf{P-Value} \\ \hline
CSP             & 66.6          & 57.8          & 61.3          & 94.1          & 80.6          & 75.0          & 72.5          & 89.4          & 85.6          & 75.9          & 12.6         & 0.0039           \\
FBCSP           & 70.0          & 60.4          & 60.9          & 97.5          & 93.1          & 80.6          & 78.1          & 92.5          & 88.9          & 80.2          & 14.0         & 0.0039           \\ 
WaSF-ConvNet    & 74.0          & 64.0          & \textbf{86.0} & 98.0          & 86.0          & 73.0          & \textbf{89.0} & 93.0          & 81.0          & 82.7          & 10.7         & 0.25             \\
EEGNet       & 77.5          & 61.1          & 63.1          & \textbf{98.4} & \textbf{96.6} & 83.8          & 84.4          & 92.8          & 88.4          & 82.9          & 13.5         & 0.02             \\
ConvNet      & 74.4          & 56.1          & 57.5          & 97.5          & 95.3          & 82.2          & 79.7          & 87.5          & 86.6          & 79.6          & 14.8         & 0.0039           \\
LMDA-Net            & 81.2          & 62.1          & 71.8          & \textbf{98.4} & 95.6          & 89.3          & 85.0          & \textbf{94.6} & 91.8          & 85.5          & 12.0         & 0.25             \\ \hline
TSFF-img  & 76.6          & 60.7          & 55.9          & 96.9          & 84.1          & 76.6          & 78.4          & 93.4          & 84.4          & 78.5          & 13.5         & 0.0039           \\ 
TSFF-Net & \textbf{83.8} & \textbf{66.8} & 67.8          & \textbf{98.4} & 95.6          & \textbf{90.0} & 86.2          & 93.8          & \textbf{95.0} & \textbf{86.4} & 11.8         & -                \\ \hline
\end{tabular}
    \begin{tablenotes}
        \footnotesize
        \item P-value denotes the Wilcoxon signed-rank test between TSFF-Net and other results.
        \item The results of CSP, ConvNet and EEGNet in this dataset are derived from \cite{tang2020motor}, \cite{zhao2020deep} and \cite{miao2022priming}).
    \end{tablenotes}
\end{threeparttable}
\end{table*}

\subsubsection{Quadruple classification of BCI4-2A}
To the best of our knowledge, no work has been done to study the problem of motor imagery multi-categorization in low-channel scenarios, likely because it is challenging to cover the brain regions corresponding to the relevant neural activity in such scenarios. However, due to the volume conduction effect, the electrical signals measured by a single EEG electrode result from superimposed neural activity. This leads to a decrease in the spatial resolution of the EEG but also provides a possibility for motor imagery multi-classification tasks in low-channel scenarios.

To investigate this, we validated the four-category motor imagery performance of the model using C3, Cz, and C4 channels based on the source code provided by EEGNet [\cite{lawhern2018eegnet}], ConvNet [\cite{schirrmeister2017deep}], and LMDA-Net [\cite{miao2023lmda}]. 
The experimental results showed that the time-series based models still outperformed the time-frequency spectrum based model in terms of classification performance. ConvNet surpassed LMDA-Net and TSFF-img in classification performance under various scenarios. However, by fusing the features learned by the time-space feature extraction network and time-frequency feature extraction network, TSFF-Net outperformed ConvNet in terms of the majority of participants and average accuracy. This further supports the effectiveness of the time-space-frequency fusion framework presented in this paper, indicating that the time-space-frequency fusion framework is more advantageous in low-channel motor imagery scenarios.
\begin{table*}[]
\caption{4-class classification performance of different algorithms on BCI4-2A}
\centering
\label{tabel:4}
\begin{threeparttable}
\begin{tabular}{ccccccccccccc}
\hline
\textbf{Methods}            & \textbf{A01}  & \textbf{A02}  & \textbf{A03}  & \textbf{A04}  & \textbf{A05}  & \textbf{A06}  & \textbf{A07}  & \textbf{A08}  & \textbf{A09}  & \textbf{Mean}  & \textbf{Std} & \textbf{P-Value} \\ \hline
LMDA-Net             & 69.8          & 52.8          & 74.0          & 50.7          & 48.3          & 43.8          & 86.1          & 54.2          & 72.9          & 61.4          & 14.6         & 0.011            \\
EEGNet               & 68.4          & 50.4          & 77.8          & 47.2          & 43.4          & 45.8          & \textbf{86.8} & 55.9          & 69.8          & 60.6          & 15.6         & 0.007            \\
ConvNet              & 67.7          & \textbf{58.7} & 72.9          & \textbf{52.1} & \textbf{59.4} & 43.1          & 86.8          & 56.6          & 73.3          & 63.4          & 13.1         & 0.49             \\ \hline
TSFF-img       & 66.7          & 58.3          & 74.3          & 45.1          & 47.6          & 39.9          & 75.0          & 48.3          & 69.8          & 58.3          & 13.5         & 0.007            \\ 
TSFF-Net & \textbf{74.3} & 56.3          & \textbf{78.8} & 51.7          & 52.1          & \textbf{50.3} & 86.1          & \textbf{61.5} & \textbf{75.7} & \textbf{65.2} & 13.6         & -                \\ \hline
\end{tabular}
    \begin{tablenotes}
        \footnotesize
        \item P-value denotes the Wilcoxon signed-rank test between TSFF-Net and other results.
        \item The results of EEGNet, ConvNet, and LMDA-Net were obtained using their open-source network architectures, with an experimental setup identical to that of TSFF-Net.
    \end{tablenotes}
\end{threeparttable}
\end{table*}

\subsection{Comparison of ANNs and CSP related methods} \label{sec:ANNs and CSP}
\begin{figure*}
\centerline{\includegraphics[width=0.75\textwidth]{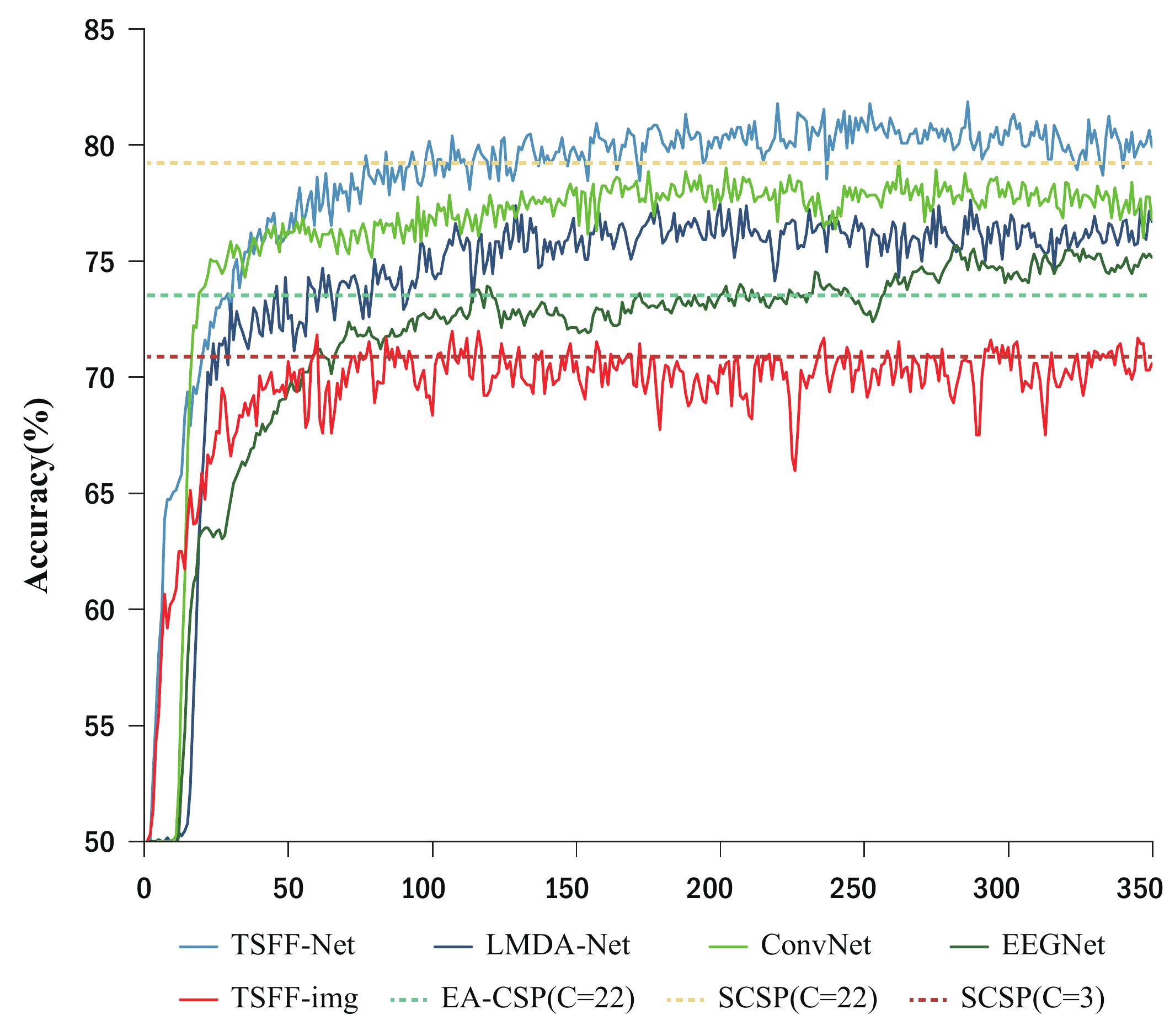}}
\caption{Comparing Binary Classification performance of ANNs and CSP-Based Methods on BCI4-2A.During each training epoch, ANNs calculate the average accuracy across all participants, while CSP-based methods obtain the best accuracy for each participant first and then take the average.}
\label{fig:2}
\end{figure*}

Neural network models require a certain number of training epochs to converge. Therefore, we compared the performance of EEGNet, ConvNet, LMDA, TSFF-img, and TSFF-Net in the BCI4-2A binary classification task over 350 epochs, using the SCSP(3channels), SCSP(22channels), and EA-CSP(22channels) as benchmarks. It is important to note that Figure \ref{fig:2} displays the average accuracy of the neural network model for each training epoch for all participants. Since the convergence rate may vary slightly among participants, taking the average accuracy of all participants for each epoch may lead to a decrease in the mean accuracy of the neural network model. Consequently, the average accuracy of the neural network shown in Figure \ref{fig:2} is lower than the average accuracy of all participants presented in Table \ref{tabel:2a2cls}.
In contrast, CSP-related methods compute the best classification result for each participant before averaging them to obtain the final result.

As shown in Figure \ref{fig:2}, all the neural network models reach convergence after less than 150 epochs of training. The accuracy of all time-series-based models after convergence exceeds that of SCSP(3channels). After approximately 260 training epochs, EEGNet, ConvNet, and LMDA-Net (TSFF-raw) all surpass the classification performance of EA-CSP(22). This demonstrates the advantage of time-series models in low-channel motor imagery scenarios.

From Figure \ref{fig:2}, it can be observed that the classification performance of the proposed TSFF-Net exceeds that of the SCSP(22channels) in the scenario using only three channels. This suggests that the proposed TSFF-Net can effectively address the limitations of feature extraction in low-channel scenarios, thus providing more convenience for the use of low-channel MI-EEG.

\subsection{Comparison of time-frequency spectrogram based models} \label{sec:spectrogram models}
In this subsection, we conduct a detailed evaluation of the feature extraction capabilities of different models for time-frequency spectrograms in low-channel scenarios. Specifically, we compare the classification performance of the TSFF-img method with that of classical natural image convolutional networks, and explore the impact of different stitching methods and downsampling sizes on the performance of the TSFF-img network.

\subsubsection{Comparison of TSFF-img and classical convolutional neural network}

\cite{khare2020time} employed pre-trained AlexNet, VGG, and ResNet networks to recognize time-frequency spectrograms and subsequently identify emotion-related information embedded in EEG signals. To the best of our knowledge, there is currently no research on motor imagery classification based on time-frequency spectrograms in the context of low-channel EEG scenarios. Therefore, we compare the performance of TSFF-img with pre-trained AlexNet, VGG16, and ResNet18 for motor imagery classification.

\begin{figure*}
\centerline{\includegraphics[width=0.9\textwidth]{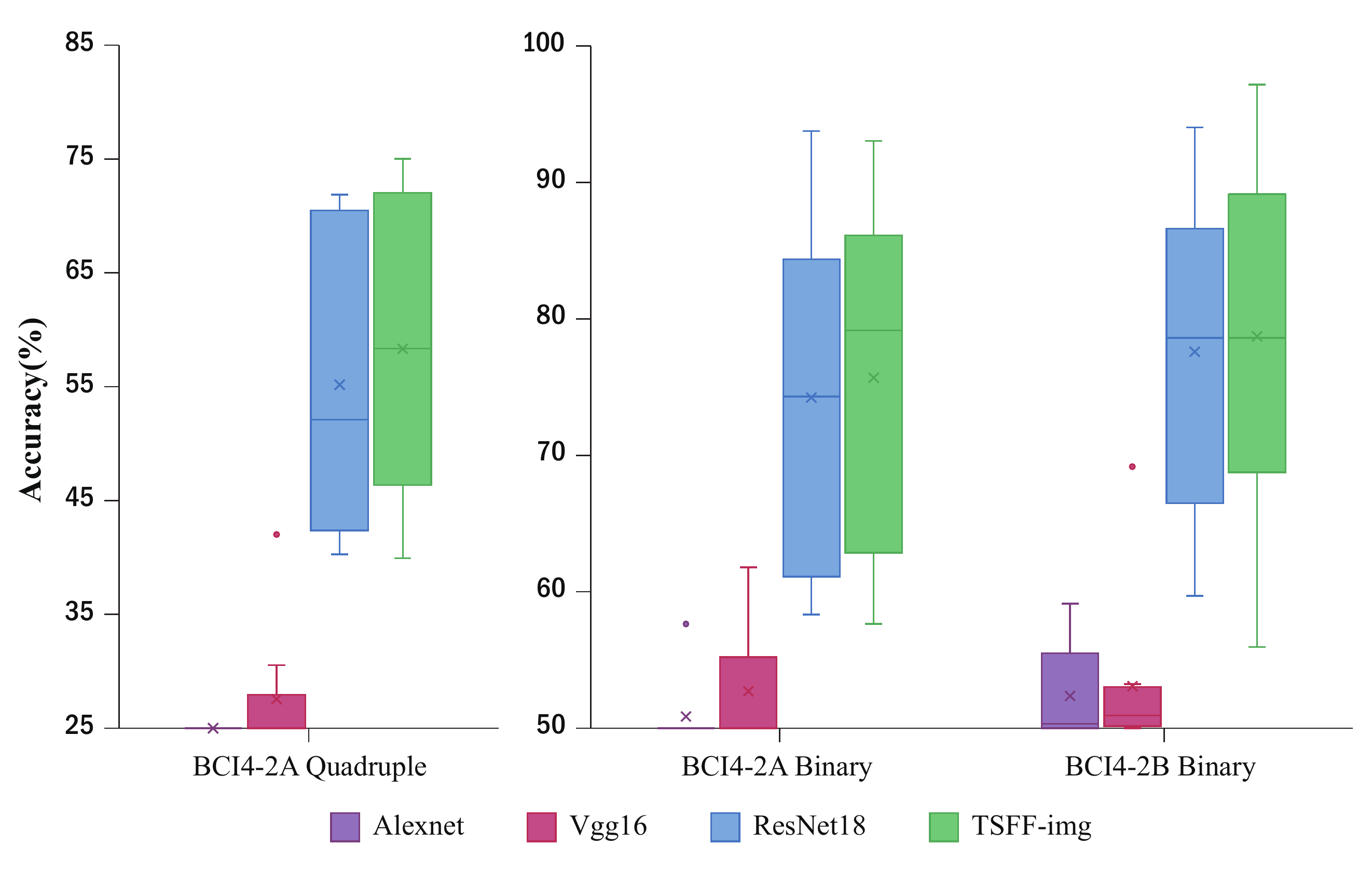}}
\caption{Comparative performance of TSFF-img and pre-trained CNNs in low-channel EEG-based motor imagery classification}
\label{fig:3}
\end{figure*}

To provide a detailed illustration of the performance of various models on different datasets, we utilized boxplots to depict the classification performance of each model.
As illustrated in Figure \ref{fig:3}, the pre-trained AlexNet and VGG16 do not exhibit superior performance in either binary or quadruple classification tasks. Both pre-trained AlexNet and VGG16 perform at levels close to random guesses under the same training conditions. In contrast, TSFF-img, with its shallow and lightweight network architecture, outperforms AlexNet and VGG16. When compared with ResNet18, TSFF-img surpasses ResNet18 in terms of mean and median indices for the aforementioned motor imagery tasks. These results suggest that, compared to pre-trained ResNet18, AlexNet, and VGG16, TSFF-img demonstrates a distinct advantage in the classification performance of motor imagery time-frequency spectrograms.

We also investigated the impact of stitching and downsampling sizes on the classification accuracy of time-frequency spectrograms from different channels in the BCI4-2B dataset. The dimensions of the time-frequency spectrograms for each channel are [3, W, L]. We experimented with stitching the time-frequency spectrograms in various dimensions, such as [3, W, 3L] for lengthwise stitching, [3, 3W, L] for widthwise stitching, and [9, W, L] for depthwise stitching. Following Khare et al. \cite{khare2020time}, we then downsampled the stitched time-frequency maps to [3, 224, 224].

Table \ref{tabel:5} reveals that widthwise stitching of time-frequency spectrograms from different channels yields better classification performance than lengthwise and depthwise stitching. Consequently, we selected widthwise stitching for TSFF-img in our experiments. Furthermore, we assessed the influence of varying downsampling sizes on the classification results in widthwise stitching. Table \ref{tabel:5} indicates that downsampling to 224 provides a higher average classification accuracy compared to downsampling to 72, 128, and 512. Thus, we chose a downsampling size of 224 for subsequent time-frequency spectrogram analyses.

Considering the results shown in Table \ref{tabel:5}, we opted to stitch time-frequency spectrograms from different channels widthwise and then downsample them to [3, 224, 224] before inputting them into the TSFF-img for training (as illustrated in Figure \ref{fig:tsff}).

\begin{table*}[]
\caption{Effects of different splicing methods and downsampling sizes of TSFF-img on the classification results of BCI4-2B}
\centering
\label{tabel:5}
\begin{threeparttable}
\begin{tabular}{ccccccccccccc}
\hline
\textbf{Methods}             & \textbf{B01}  & \textbf{B02}  & \textbf{B03}  & \textbf{B04}  & \textbf{B05}  & \textbf{B06}  & \textbf{B07}  & \textbf{B08}  & \textbf{B09}  & \textbf{AVE}  & \textbf{STD} & \textbf{P-Value} \\ \hline
TSFF-img(L224) & 74.7          & 60.0          & \textbf{56.3} & 95.9          & 82.8          & 73.8          & 79.4          & 92.5          & 81.6          & 77.4          & 13.2         & 0.054            \\
TSFF-img(D224)         & 75.0          & 56.1          & 55.6          & 95.6          & 82.5          & 75.3          & \textbf{80.9} & 91.3          & 81.3          & 77.1          & 13.8         & 0.07             \\
TSFF-img(W72)               & 70.3          & 58.2          & 57.2          & 95.6          & 71.3          & 70.6          & 70.3          & 91.3          & 78.4          & 73.7          & 13.1         & 0.01             \\
TSFF-img(W128)              & 74.1          & 54.6          & 57.8          & 95.6          & 80.0          & 74.7          & 78.4          & 92.2          & 81.3          & 76.5          & 13.6         & 0.049            \\
TSFF-img(W512)              & 73.8          & 57.5          & 55.9          & 95.3          & \textbf{86.6} & \textbf{78.1} & 77.8          & 92.8          & 81.3          & 77.7          & 13.8         & 0.18             \\ \hline
TSFF-img(W224)  & \textbf{76.6} & \textbf{60.7} & 55.9          & \textbf{96.9} & 84.1          & 76.6          & 78.4          & \textbf{93.4} & \textbf{84.4} & \textbf{78.5} & 13.5         & -                \\ \hline
\end{tabular}
    \begin{tablenotes}
        \footnotesize
        \item P-value denotes the Wilcoxon signed-rank test of SEPCONWIDCAT224 and other results.
        \item The notation (L224) denotes the operation of lengthwise stitching of time-frequency spectrograms, followed by downsampling to [3, 224, 224]. Similarly, the notations W and D represent the operations of widthwise stitching and depthwise stitching, respectively.
    \end{tablenotes}
\end{threeparttable}
\end{table*}

\subsection{Ablation study} \label{sec:ablation}

\begin{figure*}
\centerline{\includegraphics[width=0.75\textwidth]{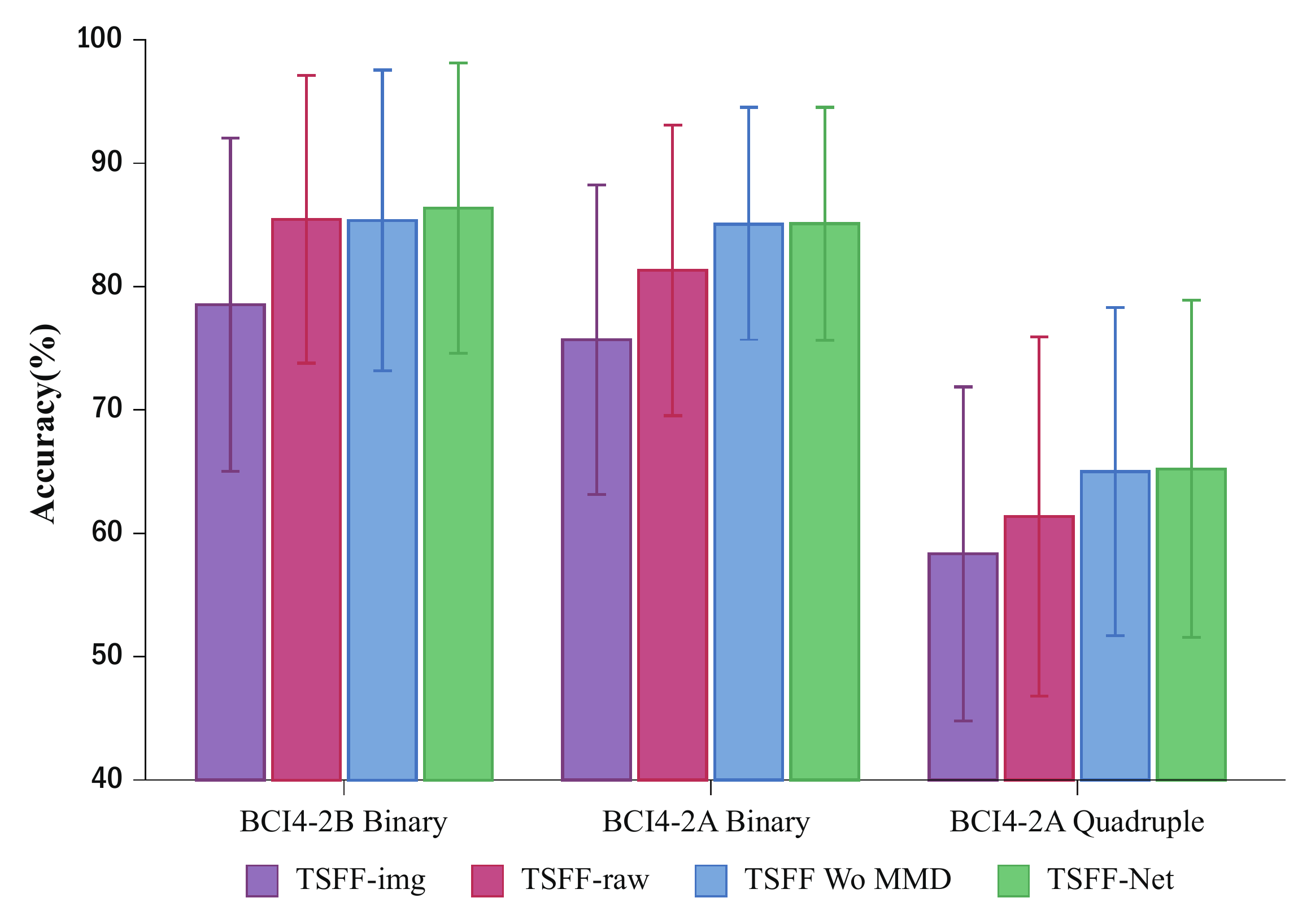}}
\caption{Ablation study of TSFF-Net on classification performance in low-channel scenarios}
\label{fig:4}
\end{figure*}

The proposed fusion method comprises a time-space feature extraction network (TSFF-raw), a time-frequency feature extraction network (TSFF-img), and a feature fusion and classification component. The ablation experiment results are presented in Figure \ref{fig:4}, which depict the MI classification performance in different scenarios using only the TSFF-raw network, only the TSFF-img network, feature fusion without MMD loss, and the complete proposed fusion framework (TSFF-Net). The histogram in Figure \ref{fig:4} displays the average accuracy across all participants, while the error bars represent the standard deviation of the classification results for different participants.

From Figure \ref{fig:4}, it is evident that the time-series based method (TSFF-raw) demonstrates superior classification performance compared to the time-frequency spectrogram based method in the same test scenarios. Consequently, the proposed fusion method compensates for the time-frequency features neglected in the TSFF-raw network during feature extraction through the TSFF-img network. In the feature fusion stage, the classification results from the BCI4-2A dataset reveal that the weighted fusion of time-space features and time-frequency features significantly enhances classification accuracy. When the weight of the time-frequency feature set was adjusted to 0.001 (considering the differing orders of magnitude between time-frequency spectrograms and time-series EEG data when input into the neural network), BCI4-2A achieves the highest average accuracy of 85.1\% for binary classification of left- and right-handed tasks and 65.0\% for quadruple classification tasks.

However, for the BCI4-2A binary classification task, employing MMD loss does not contribute to improving average accuracy. By adding MMD loss and setting the weight of MMD loss to 0.1 (with the corresponding weight of time-frequency features at 0.01), the average accuracy of the BCI4-2A quadruple classification task reaches 65.2\%. For BCI4-2B, using only weighted fusion of time-space features and time-frequency features is insufficient for enhancing average accuracy. By setting the weight (i.e. the $\lambda$ above) of MMD loss to 1, with the corresponding weight of time-frequency features at 0.001, the average accuracy of the BCI4-2B dataset reaches 86.4\%. To the best of our knowledge, this is also the highest average accuracy result under the same test conditions on BCI4-2B.

By comparing the experimental results presented above, it is evident that the proposed TSFF-Net can effectively integrate time-space and time-frequency features to compensate for the lack of single modality features in low-channel motor imagery classification. This approach provides a powerful classification algorithm for MI-EEG data collected in low-channel scenarios, thereby promoting the widespread use of such devices in various applications.

\section{Conclusion}
\label{sec:Conclusion}

In this paper, we propose TSFF-Net, a novel network architecture that effectively addresses the challenges in decoding low-channel motor imagery by integrating time-space-frequency features. TSFF-Net compensates for the limitations of single-mode feature extraction networks based on time-series or time-frequency modalities.  By fusing time-space-frequency features, TSFF-Net outperforms other state-of-the-art methods in classification accuracy on both BCI4-2A and BCI4-2B dataset. Specifically, in the BCI4-2A dataset, TSFF-Net achieves superior performance with only three channels, surpassing some classical algorithms designed based on 22 channels. 
Furthermore, as few studies have explored the classification of motor imagery based on three-channel time-frequency spectrograms, we propose a lightweight and shallow decoding architecture, TSFF-img, for feature extraction. Our experimental results demonstrate that TSFF-img outperforms popular natural image network architectures such as AlexNet, VGG, and ResNet for feature extraction from EEG time-frequency spectrograms. This suggests that lightweight shallow neural network architectures are also suitable for feature extraction from time-frequency spectrograms. In addition, comprehensive experiments are designed to validate our proposed approach, and the statistical analysis of experimental results is presented.
Overall, TSFF-Net holds significant potential in decoding low-channel motor imagery, providing valuable insights for algorithmically enhancing low-channel EEG decoding. The success of TSFF-Net opens up new avenues for further research and applications, particularly in portable and entertainment settings where low-channel EEG devices are of paramount importance.

\section*{Statistics and reproducibility}  
We used the hold-out test set method in all experiments and fixed all initialization seeds to ensure the reproducibility of the results obtained from the neural network model.
The hyperparameters used in this study were not optimized for each participant individually, but were instead set based on the average accuracy of all participants. Specifically, the same initialization parameter was applied to all participants for each classification scenario.
Given the large individual differences observed in our study, we computed the average accuracy and variance for each participant. Additionally, we conducted a Wilcoxon signed-rank test to determine the statistical significance (i.e., p-value) of the proposed method in comparison to other approaches.

\section*{Data and code availability statement}
The raw EEG data that support the findings of this study are available in \url{www.bbci.de/competition/iv/\#dataset2a} and \url{www.bbci.de/competition/iv/\#dataset2b}, respectively.  

The source code for EEGNet , ConvNet and LMDA-Net is publically available at the following webpage: \url{https://github.com/vlawhern/arl-eegmodels}, \url{https://github.com/TNTLFreiburg/braindecode} and \url{https://github.com/MiaoZhengQing/LMDA-Code}, respectively. We are pleased to provide publiclly source code for TSFF-Net at \url{https://github.com/MiaoZhengQing/TSFF}.

\section*{Credit authorship contribution statement}
\textbf{Zhengqing Miao}: Conceptualization, Methodology, Formal analysis, Investigation, Visualization, Writing -original draft  \textbf{Meirong Zhao} Supervision, Funding acquisition. 

\section*{Competing Interests statement}
The authors declare no competing interests. 

\section*{Acknowledgments}
We extend our gratitude to Yiwei Yang for polishing the figures. We would also like to express our heartfelt appreciation to the anonymous reviewers for their constructive feedback and valuable suggestions, which significantly contributed to the improvement of the quality of our paper.

\bibliographystyle{cas-model2-names}
\bibliography{TSFF_references}

\end{document}